\documentclass{article}

    \PassOptionsToPackage{numbers, compress}{natbib}

\usepackage[final]{neurips_2023}

\usepackage[utf8]{inputenc} 
\usepackage[T1]{fontenc}    
\usepackage{hyperref}       
\usepackage{url}            
\usepackage{booktabs}       
\usepackage{amsfonts}       
\usepackage{nicefrac}       
\usepackage{microtype}      
\usepackage{xcolor}         
\usepackage{mathtools}      
\usepackage{color, colortbl} 
\definecolor{Gray}{gray}{0.95}
\usepackage{caption,subcaption}
\usepackage{floatrow}
\newfloatcommand{capbtabbox}{table}[][\FBwidth]

\title{Physics-Informed Calibration of Aeromagnetic Compensation in Magnetic Navigation Systems using Liquid Time-Constant Networks}

\author{%
  Favour Nerrise\thanks{F.N. is also an AI Ph.D. Resident at SandboxAQ, alternative email: favour.nerrise@sandboxaq.com.} \\
  Department of Electrical Engineering \\
  Stanford University, CA, USA\\
\texttt{fnerrise@stanford.edu} \\
  \And
  Andrew (Sosa) Sosanya \\
  SandboxAQ \\
  Palo Alto, CA, USA \\
\texttt{andrew.sosanya@sandboxaq.com } \\
  \And
  Patrick Neary \\
  SandboxAQ \\
  Palo Alto, CA, USA \\
\texttt{patrick.neary@sandboxaq.com } \\
}

\begin{document}

\maketitle

\begin{abstract}
    Magnetic navigation (MagNav) is a rising alternative to the Global Positioning System (GPS) and has proven useful for aircraft navigation. Traditional aircraft navigation systems, while effective, face limitations in precision and reliability in certain environments and against attacks. Airborne MagNav leverages the Earth's magnetic field to provide accurate positional information. However, external magnetic fields induced by aircraft electronics and Earth's large-scale magnetic fields disrupt the weaker signal of interest. We introduce a physics-informed approach using Tolles-Lawson coefficients for compensation and Liquid Time-Constant Networks (LTCs) to remove complex, noisy signals derived from the aircraft’s magnetic sources. Using real flight data with magnetometer measurements and aircraft measurements, we observe up to a 64\% reduction in aeromagnetic compensation error (RMSE nT), outperforming conventional models. This significant improvement underscores the potential of a physics-informed, machine learning approach for extracting clean, reliable, and accurate magnetic signals for MagNav positional estimation.

\end{abstract}

\section{Introduction}\label{intro}
In recent years, magnetic anomaly navigation (MagNav) has proven to be a viable fallback to the Global Positioning System (GPS) for aircraft navigation due to its reliability against jamming\footnote{GPS jamming disrupts a GPS receiver by transmitting powerful signals that block or interfere with the original GPS signal\cite{westbrook2019global}.} or spoofing \footnote{GPS spoofing manipulates existing GPS signals to mislead a receiver about its position, velocity, and timing\cite{psiaki2016gnss}.} attacks, independence from atmospheric weather conditions, and unsupervised position estimation \cite{9506809}. Airborne MagNav estimates positioning by correlating aircraft magnetometer readings to anomaly maps of the Earth’s crustal magnetic field \cite{9506809}. Successful airborne MagNav studies have shown high accuracies in navigation estimation, with up to 10m distance root-mean-squared errors at low altitudes and tens of meters to kilometers at varying altitudes above ground level \cite{canciani2017airborne, wilson2006passive, lee2020magslam}. This is only slightly inferior to GPS estimation for aircraft navigation, which is accurate down to seven meters ($\sim23$ feet), 95\% of the time, anywhere on or near the Earth's surface \cite{FAAGPS}. However, GPS signals are increasingly unreliable due to their dependence on a network of vulnerable satellites. Additionally, GPS signals are prone to initialization delays and dropouts. MagNav signals, in contrast, constantly emit from the invisible Earth’s geomagnetic field, are well-defined, and are mappable. 

However, there are differences in MagNav estimation accuracies, which can be attributed to the altitude flown, map quality, and calibration of the platform’s magnetic field. Additionally, magnetic measurements derived from scalar and vector magnetometers (magnetic sensors affixed to the aircraft) depend on a nonlinear combination of the Earth’s magnetic field and the aircraft’s magnetic field, which can cause a navigation estimate that drifts over time due to corrupted magnetic fields (namely, the permanent field, the induced field, and the eddy-current field, as defined by Tolles and Lawson \cite{tolles1950magnetic}). Furthermore, time-varying magnetic fields from sources like electronic equipment, communication systems, and temperature-dependent magnetic properties of sensors cause deterministic and stochastic effects that violate static assumptions of the classical Tolles–Lawson (T-L) error sources used for predicting aircraft magnetic interference \cite{leliak1961identification}. For a real-time application, a model continuously estimates and eliminates the unknown T-L coefficients of the corrupted magnetic fields. 

In this work, we aim to reduce contamination fields and improve calibration accuracy through aeromagnetic compensation, i.e., eliminating interference related to the aircraft’s magnetic field from the total magnetometer output signal to improve magnetometer measurement harmonization to a magnetic anomaly map. Traditional state space estimation models, like Kalman filters, have previously been applied to resolve this but are limited in handling the nonlinear nature of the underlying dynamical system and are highly sensitive to perturbations \cite{9506809, beravs2014magnetometer, siebler2020localization, noriega2017recent}. Neural networks have also been used to improve the calibration of aeromagnetic compensation \cite{gnadt2022machine, laoue2023neural, xu2020deepmad, ma2018uncertainty, hezel2020improving}, but are typically traditional models that fail to leverage inherent physical laws of the system, are computationally expensive, and are not robust to high-dimensional, time-series data. We propose a two-step machine learning framework that: (1) uses a physics-informed neural network, a Liquid Time-Constant network, to capture complex, nonlinear dynamics of airborne MagNav and to detect the weak magnetic anomaly signal from a noisy environment; (2) derives the coefficients of known magnetic effects from the airplane that can be used to update a T-L model and improve calibration accuracy.

\begin{figure}
  \centering
  \includegraphics[width=0.95\textwidth]{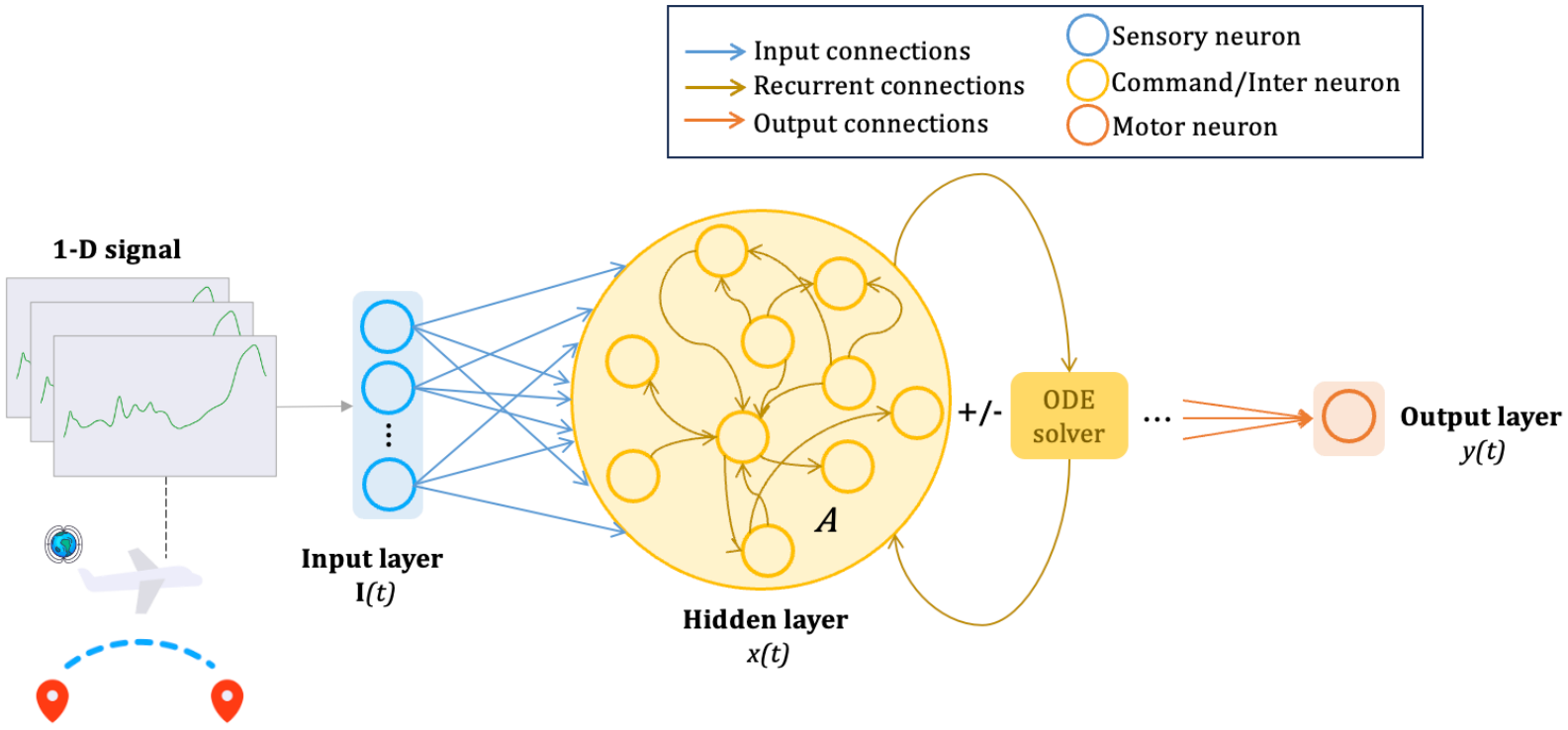}
  \caption{The architecture for the Liquid Time-Constant Network comprises the input, perception layer, a liquid layer (i.e., a time-continuous gating mechanism), and an output layer. Here, $I(t)$ is the real input signal, and $y(t)$ is the predicted, filtered signal.}
  \label{fig:LTC_model}
\end{figure}

\section{Methodology}
\label{methods}
\subsection{Liquid Time-Constant Networks}
    Liquid Time-Constant (LTC) networks are a promising variant of recurrent neural networks (RNNs) designed to capture nonlinear patterns of high-dimensional, sequential tasks \cite{hasani2021liquid}. While standard, discrete RNNs approximate continuous dynamics step-by-step via dependencies on the history of their hidden states, LTCs approximate continuous dynamics using ordinary differential equations (ODEs) \cite{chen2018neural} with more elaborate temporal steps. Additionally, their timing behavior is adaptive to the input (how fast or slow they respond to some stimulus can depend on the specific input). These ODE-based neural networks, however, are slower in training and inference due to using advanced solvers. This lag intensifies with increasing data complexity, especially for tasks like autonomous driving, financial time series, and physics simulations. In this work, we leverage an LTC model enhanced by a closed-form continuous-depth solution (CfC) \cite{hasani2022closed} that combines the rich modeling capabilities of ODE-based models without needing a solver, resulting in significantly faster training and inference speeds. This model has been applied to various sequence modeling problems like robot and drone navigation \cite{xiao2023barriernet, chahine2023robust}, vision-based autonomous driving \cite{hasani2022closed, xiao2023barriernet}, and forecasting sea ice concentration \cite{beveridge2022interpretable} and cardiac abnormality detection \cite{huang2023efficient}. Given a 1-D, time-series input signal with $m$-features, $\mathbf{I}(t)^{m\times1}$, at time-step $t$, a hidden state vector with $D$ hidden units, $\mathbf{x}^{(D\times1)}(t)$, and a time-constant parameter vector, $w_{\tau}^{(D)}$, the hidden state of an LTC network is determined by the solution to the following problem (see Fig. \ref{fig:LTC_model}):
    \begin{equation}\label{eq:ltc}
        \frac{d\mathbf{x}}{dt} = w_{\tau} + f(\mathbf{x}, \mathbf{I}, \theta)\mathbf{x}(t) + A f (\mathbf{x}, \mathbf{I}, \theta)
    \end{equation}
    where $A$ is a bias vector, and $f$ is a neural network parametrized by $\theta$.
    
    \subsection{Neural Circuit Policies (NCPs)}
    A Neural Circuit Policy network \cite{lechner2020neural} is a sparse recurrent neural network modeled after nervous system of the C. elegans nematode \cite{corsi2015transparent}. It has four recurrent connection types: sensory, inter-neurons, command, and motor neurons. Sensory neurons receive environmental observations (input, $x$), which are forwarded to inter-neurons. Command neurons, linked to inter-neurons, handle decision-making. Motor neurons, connected to command neurons, manage actuation (output, $y$). Synapses can be excitatory or inhibitory and affect neuron activation. Unlike traditional neural networks, each neuron’s dynamics are governed by an ODE, which allows for complex dynamics. We train these networks using NCP wirings and the Proximal Policy Optimization (PPO) method \cite{schulman2017proximal}.

    \subsection{Closed-form Continuous-time (CfC)
Neural Networks}
   A CfC network is a recurrent neural network that resolves a bottleneck requiring a numerical ODE solver by approximating the closed-form solution of the differential equation. Accordingly, CfCs deliver higher efficiency and achieve adaptive, causal, and continuous-time attributes with faster training and inference time. The hidden state of the CfC model can be explicitly written as: 
    \begin{equation} \label{eq:cfc_fin}
        \begin{aligned}
        \mathbf{x}(t) = \underbrace{\sigma(- f (\mathbf{x}, \mathbf{I}; \theta_{f} ) \mathbf{t})}_\text{time-continuous gating} \odot g(\mathbf{x}, \mathbf{I}; \theta_{g}) + \\
        \underbrace{[1 - \sigma(- [f (\mathbf{x}, \mathbf{I}; \theta_{f})] \mathbf{t})]}_\text{time-continuous gating} \odot h(\mathbf{x}, \mathbf{I}; \theta_{h})
        \end{aligned}
    \end{equation}
    where $f$, $g$, and $h$ are NCP network heads and $\odot$ is the Hadamard (element-wise) product. For time-variant datasets, $t$ is set based on its timestamp or order, while for sequential applications where sample timing is irrelevant, $t$ is sampled uniformly between two hyperparameters.

\subsection{Dataset \& Pre-processing} 
    \begin{figure}[t!]
    \centering
    \begin{subfigure}[ht]{0.55\linewidth}
        \includegraphics[width=\linewidth]{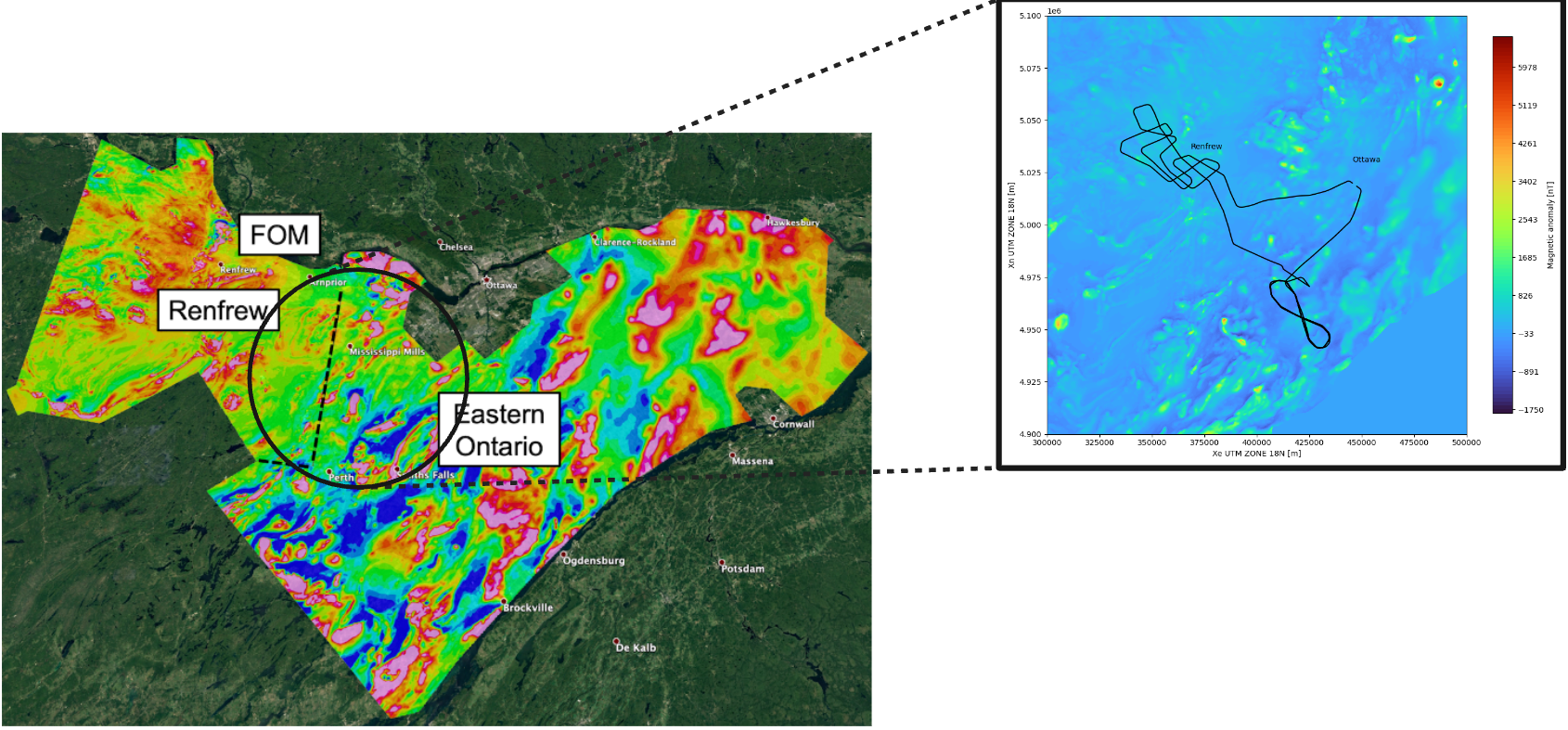}
    \end{subfigure}
    \hfill
    \begin{subfigure}[ht]{0.4\linewidth}
        \includegraphics[width=\linewidth]{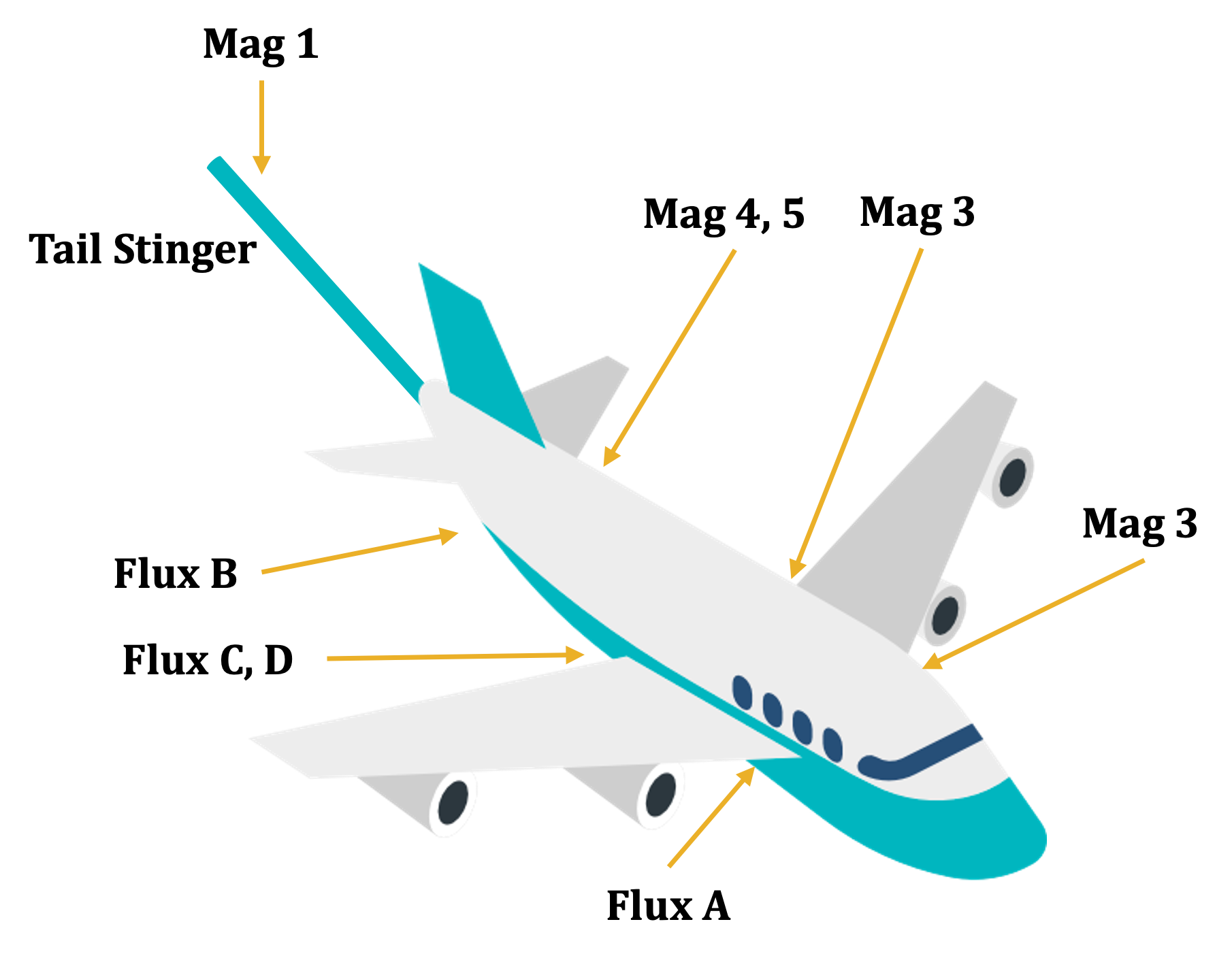}
    \end{subfigure}
    \caption{a) Flight trajectory on the magnetic anomaly of Renfrew, Canada, and Ontario, Canada, at 300m above the WGS84 ellipsoid (flight 1007). b) Locations of magnetometers positioned around the MagNav Challenge aircraft.}
      \label{fig:featureselect}
    \end{figure}
    We use the United States Air Force (USAF)-MIT Signal Enhancement for Magnetic Navigation Challenge dataset \cite{gnadt2020signal}, which aims to identify optimal aeromagnetic compensation for airborne MagNav. This dataset includes time series from six flights and captures magnetic field data from five scalar and three vector magnetometers positioned differently within the aircraft (see Fig. \ref{fig:featureselect}b). Each flight has 65 features, with some segments having INS, avionics, electrical, and radar readings. Post T-L calibration, the scalar magnetometer at the aircraft’s tail stinger, Mag 1, serves as our ground truth target signal, $\hat{y}$. Signals are sampled at 10 Hz. We apply T-L compensation to magnetometer data, subtract the International Geomagnetic Reference Field to degrade variations caused by the Earth’s core field, and correct for diurnal effects.
    
\subsection{Feature Selection}
    We standardize all features using $z$-score normalization to have zero mean and unit standard deviation. To gather more information from magnetometer measurements, we residualize magnetometers within the aircraft against the magnetometer outside the aircraft (Mag 1). To reduce overfitting and enhance the interpretability of salient features towards model performance, we apply three feature selection techniques: Spearman correlation \cite{spearman1904proof}, LASSO regularization \cite{tibshirani1996regression}, and XGBoost \cite{chen2016xgboost}, to sub-select the most important features from the dataset. Accordingly, we select the following features: magnetometer 4 \& 5 measurements, aircraft positioning and orientation, INS velocities, altitude, and several electrical measurements (V\_BAT1, V\_BAT2, CUR\_ACLo, CUR\_FLAP, CUR\_TANK, CUR\_IHTR).

\subsection{Experiment Setup}
     We treat each flight as a sample to prevent overfitting due to looping flight segments within the same flight. We apply K-fold cross-validation \cite{mosteller1968data} for resampling. We train all models on an NVIDIA Tesla T4 GPU using Python 3.10. We use PyTorch \cite{paszke2019pytorch} to build the LTC module and conduct hyperparameter search, architecture search, and tuning using Optuna \cite{akiba2019optuna}. We use an LTC and CfC cell with NCP wirings (AutoNCP), 64 inter and command neurons, and one output neuron for each trial. We optimize the model using Adam ($lr = 1e-3$), mean squared error (MSE) loss, and a dropout of $50\%$. We train for $300$ epochs with a batch size of $64$ samples. We use the Root Mean Square Error (RMSE) to measure model performance:
    {\small
        \begin{equation}\label{eq:costfuncs}
            \text{RMSE} = \sqrt{\frac{1}{N} \sum_{i=1}^{N} (y_i - \hat{y}_i)^2}
        \end{equation}
    }
where $y_i$ and $\hat{y}_i$ are the truth and predicted signals, respectively.
 
\section{Results}
\label{results}
We conducted experiments on the Liquid Time-Constant (LTC) and Liquid Time-Constant-Closed-form Continuous model (LTC-CfC). We compared LTC models against three baseline models: Multi Perceptron (MLP), Convolutional Neural Network (CNN), and Long Short-Term Memory (LSTM) Network. We also compared the performance of our proposed approach against the T-L state estimation method for classical aerocompensation calibration.

\begin{figure}[h!]
\begin{floatrow}
\capbtabbox{%
    \resizebox{.47\textwidth}{!}{
        \begin{tabular}{lll}
            \toprule
            \textbf{Model} & \textbf{Flt1003 [nT]}  & \textbf{Flt1007 [nT]} \\
            \midrule
            T-L & 58.85  & 45.13  \\
            \midrule 
            LSTM     & 41.79  & 42.18  \\
            MLP      & 30.47 & 26.23  \\
            CNN      & 26.05  & 30.56  \\
            \midrule 
            LTC     & 20.31  &   22.89 \\
            \midrule 
            \rowcolor{Gray}
            \textbf{LTC-CfC} & \textbf{18.20}  &  \textbf{19.14}   \\
            \bottomrule
          \end{tabular}
    }
}{%
  \caption{Model comparison of aerocompensation calibration error (RMSE nT) for flights 1003 and 1007.}%
  \label{tab:res_tab}
}
\ffigbox{%
  \includegraphics[width=0.95\linewidth]{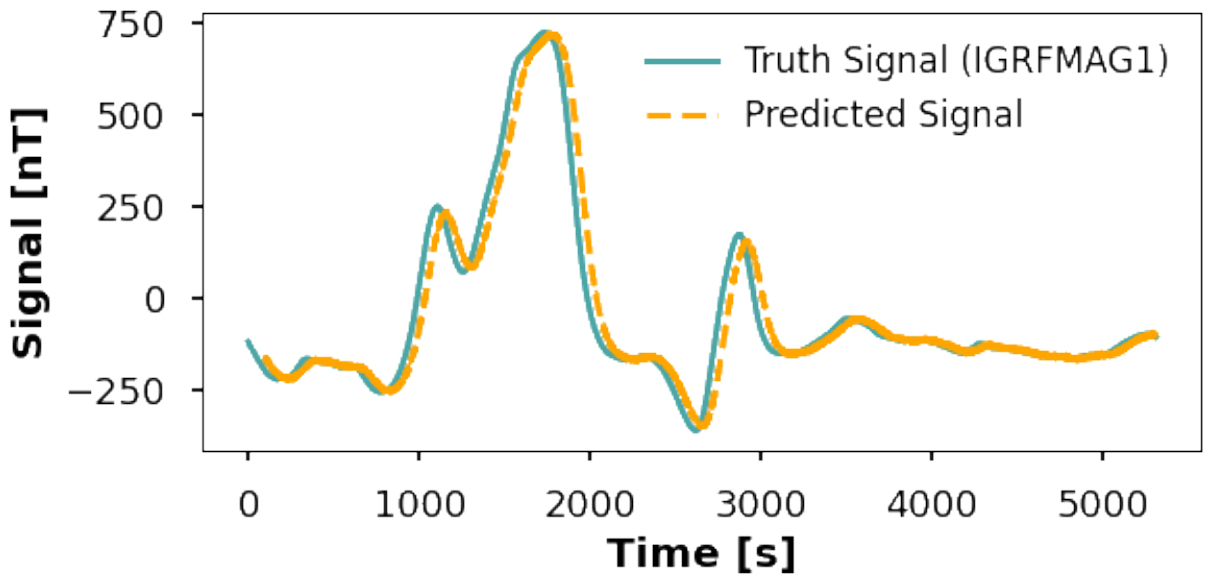}
}{%
  \caption{Truth signal (IGRF-corrected Mag 1) vs. predicted signal [nT] for a portion of flight 1007.}%
  \label{fig:real_pred}
}
\end{floatrow}
\end{figure}

As shown in Table ~\ref{tab:res_tab}, the LTC models show a significant increase in performance compared to standard non-ODE-based neural networks. The standard LTC demonstrates up to 58\% deduction in compensation error [RMSE], and the LTC-CfC similarly shows a 64\% reduction in compensation error compared to the T-L model, averaged across both test flights. Our method successfully detects weak anomaly fields with a significant accuracy threshold (see Fig. ~\ref{fig:real_pred}).

\section{Conclusion}
\label{conclusions}
This paper presents a novel, physics-informed model that leverages nonlinear dynamics in airborne MagNav systems to enhance the calibration of aeromagnetic compensation. The pipeline consists of magnetic effects corrections, Liquid Time-Constant Networks enabled with ODE-solvers or closed-form solvers, and learned coefficients for updating state estimation models for real-time use. By incorporating physics-informed learning, our approach effectively separates weak magnetic anomaly fields from disruptive aircraft magnetic noise, enabling accurate navigation for airborne MagNav. Our experimental results outperform traditional state estimation techniques. We plan on leveraging additional physics-based features learned during training and conducting external validation on new datasets. We hope to see this work applied to real-time aircraft navigation systems in the future.

\begin{ack}
This work was completed at SandboxAQ\footnote{Solving global challenges with AI + Quantum for positive impact. More at \url{sandboxaq.com}.} as a part of the Residency Program. F.N. expresses her profound gratitude to her SandboxAQ team members and peers for their dedicated expertise, mentorship, and support in completing this work. F.N. is also thankful to her research advisors at Stanford University, founding research scientists at Liquid AI, and open-source MagNav-AI collaborators and partners for their scientific and technical contributions that inspired this work.  
\end{ack}

\medskip

\bibliography{neurips_2023}

\begin{thebibliography}{34}
\providecommand{\natexlab}[1]{#1}
\providecommand{\url}[1]{\texttt{#1}}
\expandafter\ifx\csname urlstyle\endcsname\relax
  \providecommand{\doi}[1]{doi: #1}\else
  \providecommand{\doi}{doi: \begingroup \urlstyle{rm}\Url}\fi

\bibitem[Westbrook(2019)]{westbrook2019global}
Tegg Westbrook.
\newblock The global positioning system and military jamming.
\newblock \emph{Journal of Strategic Security}, 12\penalty0 (2):\penalty0 1--16, 2019.

\bibitem[Psiaki and Humphreys(2016)]{psiaki2016gnss}
Mark~L Psiaki and Todd~E Humphreys.
\newblock Gnss spoofing and detection.
\newblock \emph{Proceedings of the IEEE}, 104\penalty0 (6):\penalty0 1258--1270, 2016.

\bibitem[Canciani(2022)]{9506809}
Aaron~J. Canciani.
\newblock Magnetic navigation on an f-16 aircraft using online calibration.
\newblock \emph{IEEE Transactions on Aerospace and Electronic Systems}, 58\penalty0 (1):\penalty0 420--434, 2022.
\newblock \doi{10.1109/TAES.2021.3101567}.

\bibitem[Canciani and Raquet(2017)]{canciani2017airborne}
Aaron Canciani and John Raquet.
\newblock Airborne magnetic anomaly navigation.
\newblock \emph{IEEE Transactions on aerospace and electronic systems}, 53\penalty0 (1):\penalty0 67--80, 2017.

\bibitem[Wilson et~al.(2006)Wilson, Kline-Schoder, Kenton, Sorensen, and Clavier]{wilson2006passive}
John~M Wilson, Robert~J Kline-Schoder, Marc~A Kenton, Paul~H Sorensen, and Odile~H Clavier.
\newblock Passive navigation using local magnetic field variations.
\newblock In \emph{Proceedings of the 2006 National Technical Meeting of the Institute of Navigation}, pages 770--779, 2006.

\bibitem[Lee and Canciani(2020)]{lee2020magslam}
Taylor~N Lee and Aaron~J Canciani.
\newblock Magslam: Aerial simultaneous localization and mapping using earth's magnetic anomaly field.
\newblock \emph{Navigation}, 67\penalty0 (1):\penalty0 95--107, 2020.

\bibitem[{Federal Aviation Administration}(n.d.)]{FAAGPS}
{Federal Aviation Administration}.
\newblock How gps works, n.d.
\newblock URL \url{https://www.faa.gov/about/office_org/headquarters_offices/ato/service_units/techops/navservices/gnss/gps/howitworks}.
\newblock Accessed: November 28, 2023.

\bibitem[Tolles and Lawson(1950)]{tolles1950magnetic}
Walter~E Tolles and JD~Lawson.
\newblock Magnetic compensation of mad equipped aircraft.
\newblock \emph{Airborne Instruments Lab. Inc., Mineola, NY, Rept}, pages 201--1, 1950.

\bibitem[Leliak(1961)]{leliak1961identification}
Paul Leliak.
\newblock Identification and evaluation of magnetic-field sources of magnetic airborne detector equipped aircraft.
\newblock \emph{IRE Transactions on Aerospace and Navigational Electronics}, ANE-8\penalty0 (3):\penalty0 95--105, 1961.

\bibitem[Beravs et~al.(2014)Beravs, Begu{\v{s}}, Podobnik, and Munih]{beravs2014magnetometer}
Tadej Beravs, Samo Begu{\v{s}}, Janez Podobnik, and Marko Munih.
\newblock Magnetometer calibration using kalman filter covariance matrix for online estimation of magnetic field orientation.
\newblock \emph{IEEE transactions on instrumentation and measurement}, 63\penalty0 (8):\penalty0 2013--2020, 2014.

\bibitem[Siebler et~al.(2020)Siebler, Sand, and Hanebeck]{siebler2020localization}
Benjamin Siebler, Stephan Sand, and Uwe~D Hanebeck.
\newblock Localization with magnetic field distortions and simultaneous magnetometer calibration.
\newblock \emph{IEEE Sensors Journal}, 21\penalty0 (3):\penalty0 3388--3397, 2020.

\bibitem[Noriega(2017)]{noriega2017recent}
G~Noriega.
\newblock Recent advances in aeromagnetic compensation.
\newblock \emph{Proc. 10th Int. Congr. Prospectors Explorers}, pages 1--4, 2017.

\bibitem[Gnadt(2022)]{gnadt2022machine}
Albert Gnadt.
\newblock Machine learning-enhanced magnetic calibration for airborne magnetic anomaly navigation.
\newblock In \emph{AIAA SciTech 2022 Forum}, page 1760, 2022.

\bibitem[Laou{\'e} et~al.(2023)Laou{\'e}, Lepers, Deletraz, and Faure]{laoue2023neural}
Nathan Laou{\'e}, Arnaud Lepers, Laure Deletraz, and Charly Faure.
\newblock Neural network calibration of airborne magnetometers.
\newblock In \emph{2023 IEEE 10th International Workshop on Metrology for AeroSpace (MetroAeroSpace)}, pages 37--42. IEEE, 2023.

\bibitem[Xu et~al.(2020)Xu, Huang, Liu, and Fang]{xu2020deepmad}
Xin Xu, Ling Huang, Xiaojun Liu, and Guangyou Fang.
\newblock Deepmad: Deep learning for magnetic anomaly detection and denoising.
\newblock \emph{IEEE Access}, 8:\penalty0 121257--121266, 2020.

\bibitem[Ma et~al.(2018)Ma, Cheng, Chalup, and Zhou]{ma2018uncertainty}
Ming Ma, Defu Cheng, Stephan Chalup, and Zhijian Zhou.
\newblock Uncertainty estimation in the neural model for aeromagnetic compensation.
\newblock \emph{IEEE Geoscience and Remote Sensing Letters}, 15\penalty0 (12):\penalty0 1942--1946, 2018.

\bibitem[Hezel(2020)]{hezel2020improving}
Mitchell~C Hezel.
\newblock Improving aeromagnetic calibration using artificial neural networks.
\newblock \emph{Air Force Institute of Technology (AFIT)}, 2020.

\bibitem[Hasani et~al.(2021)Hasani, Lechner, Amini, Rus, and Grosu]{hasani2021liquid}
Ramin Hasani, Mathias Lechner, Alexander Amini, Daniela Rus, and Radu Grosu.
\newblock Liquid time-constant networks.
\newblock In \emph{Proceedings of the AAAI Conference on Artificial Intelligence}, volume~35, pages 7657--7666, 2021.

\bibitem[Chen et~al.(2018)Chen, Rubanova, Bettencourt, and Duvenaud]{chen2018neural}
Ricky~TQ Chen, Yulia Rubanova, Jesse Bettencourt, and David~K Duvenaud.
\newblock Neural ordinary differential equations.
\newblock \emph{Advances in neural information processing systems}, 31, 2018.

\bibitem[Hasani et~al.(2022)Hasani, Lechner, Amini, Liebenwein, Ray, Tschaikowski, Teschl, and Rus]{hasani2022closed}
Ramin Hasani, Mathias Lechner, Alexander Amini, Lucas Liebenwein, Aaron Ray, Max Tschaikowski, Gerald Teschl, and Daniela Rus.
\newblock Closed-form continuous-time neural networks.
\newblock \emph{Nature Machine Intelligence}, 4\penalty0 (11):\penalty0 992--1003, 2022.

\bibitem[Xiao et~al.(2023)Xiao, Wang, Hasani, Chahine, Amini, Li, and Rus]{xiao2023barriernet}
Wei Xiao, Tsun-Hsuan Wang, Ramin Hasani, Makram Chahine, Alexander Amini, Xiao Li, and Daniela Rus.
\newblock Barriernet: Differentiable control barrier functions for learning of safe robot control.
\newblock \emph{IEEE Transactions on Robotics}, 2023.

\bibitem[Chahine et~al.(2023)Chahine, Hasani, Kao, Ray, Shubert, Lechner, Amini, and Rus]{chahine2023robust}
Makram Chahine, Ramin Hasani, Patrick Kao, Aaron Ray, Ryan Shubert, Mathias Lechner, Alexander Amini, and Daniela Rus.
\newblock Robust flight navigation out of distribution with liquid neural networks.
\newblock \emph{Science Robotics}, 8\penalty0 (77):\penalty0 eadc8892, 2023.

\bibitem[Beveridge and Pereira(2022)]{beveridge2022interpretable}
Matthew Beveridge and Lucas Pereira.
\newblock Interpretable spatiotemporal forecasting of arctic sea ice concentration at seasonal lead times.
\newblock In \emph{NeurIPS 2022 Workshop on Tackling Climate Change with Machine Learning}, 2022.

\bibitem[Huang et~al.(2023)Huang, Herbozo~Contreras, Leung, Yu, Truong, Nikpour, and Kavehei]{huang2023efficient}
Zhaojing Huang, Luis~Fernando Herbozo~Contreras, Wing~Hang Leung, Leping Yu, Nhan~Duy Truong, Armin Nikpour, and Omid Kavehei.
\newblock Efficient edge-ai models for robust ecg abnormality detection on resource-constrained hardware.
\newblock \emph{medRxiv}, pages 2023--08, 2023.

\bibitem[Lechner et~al.(2020)Lechner, Hasani, Amini, Henzinger, Rus, and Grosu]{lechner2020neural}
Mathias Lechner, Ramin Hasani, Alexander Amini, Thomas~A Henzinger, Daniela Rus, and Radu Grosu.
\newblock Neural circuit policies enabling auditable autonomy.
\newblock \emph{Nature Machine Intelligence}, 2\penalty0 (10):\penalty0 642--652, 2020.

\bibitem[Corsi et~al.(2015)Corsi, Wightman, and Chalfie]{corsi2015transparent}
Ann~K Corsi, Bruce Wightman, and Martin Chalfie.
\newblock A transparent window into biology: a primer on caenorhabditis elegans.
\newblock \emph{Genetics}, 200\penalty0 (2):\penalty0 387--407, 2015.

\bibitem[Schulman et~al.(2017)Schulman, Wolski, Dhariwal, Radford, and Klimov]{schulman2017proximal}
John Schulman, Filip Wolski, Prafulla Dhariwal, Alec Radford, and Oleg Klimov.
\newblock Proximal policy optimization algorithms.
\newblock \emph{arXiv preprint arXiv:1707.06347}, 2017.

\bibitem[Gnadt et~al.(2020)Gnadt, Belarge, Canciani, Carl, Conger, Curro, Edelman, Morales, Nielsen, O'Keeffe, et~al.]{gnadt2020signal}
Albert~R Gnadt, Joseph Belarge, Aaron Canciani, Glenn Carl, Lauren Conger, Joseph Curro, Alan Edelman, Peter Morales, Aaron~P Nielsen, Michael~F O'Keeffe, et~al.
\newblock Signal enhancement for magnetic navigation challenge problem.
\newblock \emph{arXiv preprint arXiv:2007.12158}, 2020.

\bibitem[Spearman(1904)]{spearman1904proof}
C~Spearman.
\newblock The proof and measurement of association between two things.
\newblock \emph{The American Journal of Psychology}, 15\penalty0 (1):\penalty0 72--101, 1904.

\bibitem[Tibshirani(1996)]{tibshirani1996regression}
Robert Tibshirani.
\newblock Regression shrinkage and selection via the lasso.
\newblock \emph{Journal of the Royal Statistical Society Series B: Statistical Methodology}, 58\penalty0 (1):\penalty0 267--288, 1996.

\bibitem[Chen and Guestrin(2016)]{chen2016xgboost}
Tianqi Chen and Carlos Guestrin.
\newblock Xgboost: A scalable tree boosting system.
\newblock In \emph{Proceedings of the 22nd acm sigkdd international conference on knowledge discovery and data mining}, pages 785--794, 2016.

\bibitem[Mosteller and Tukey(1968)]{mosteller1968data}
Frederick Mosteller and John~W Tukey.
\newblock Data analysis, including statistics.
\newblock \emph{Handbook of social psychology}, 2:\penalty0 80--203, 1968.

\bibitem[Paszke et~al.(2019)Paszke, Gross, Massa, Lerer, Bradbury, Chanan, Killeen, Lin, Gimelshein, Antiga, et~al.]{paszke2019pytorch}
Adam Paszke, Sam Gross, Francisco Massa, Adam Lerer, James Bradbury, Gregory Chanan, Trevor Killeen, Zeming Lin, Natalia Gimelshein, Luca Antiga, et~al.
\newblock Pytorch: An imperative style, high-performance deep learning library.
\newblock \emph{Advances in neural information processing systems}, 32, 2019.

\bibitem[Akiba et~al.(2019)Akiba, Sano, Yanase, Ohta, and Koyama]{akiba2019optuna}
Takuya Akiba, Shotaro Sano, Toshihiko Yanase, Takeru Ohta, and Masanori Koyama.
\newblock Optuna: A next-generation hyperparameter optimization framework.
\newblock In \emph{Proceedings of the 25th ACM SIGKDD international conference on knowledge discovery \& data mining}, pages 2623--2631, 2019.

\end{thebibliography}


\end{document}